# Image Retrieval using Histogram Factorization and Contextual Similarity Learning

Liu Liang *, †

* College of Information Science and Engineering, Ocean University of China, Qingdao 266003, China
† State Key Laboratory for Novel Software Technology, Nanjing University, P.R. China

*Abstract*—Image retrieval has been a top topic in the field of both computer vision and machine learning for a long time. Content based image retrieval, which tries to retrieve images from a database visually similar to a query image, has attracted much attention. Two most important issues of image retrieval are the representation and ranking of the images. Recently, bag-of-words based method has shown its power as a representation method. Moreover, nonnegative matrix factorization is also a popular way to represent the data samples. In addition, contextual similarity learning has also been studied and proven to be an effective method for the ranking problem. However, these technologies have never been used together. In this paper, we developed an effective image retrieval system by representing each image using the bag-of-words method as histograms, and then apply the nonnegative matrix factorization to factorize the histograms, and finally learn the ranking score using the contextual similarity learning method. The proposed novel system is evaluated on a large scale image database and the effectiveness is shown.

*Keywords-image retrieval; bag-of-words; nonnegative matrix factorization; contextual similarity learning*

## I. Introduction

Content based image retrieval [1,2] problem is defined as follows: given a database with a large number of images, and a query image, we want to find some images from the database which is visually similar to the query. To do this, we usually first extract visual features from each image, and then compare the features of the query against each of the database images. A similarity measure will be used to measure the similarity between the query and each database image. According to the similarities between the query and the database images, the most similar images will be returned as retrieval results. There are many applications of image retrieval. For example, the google and Baidu search engine provide the image search service, which allows the user to upload and query image, and a list of similar images will be returned. Another example is medical image retrieval, which is of great help to clinic doctors.

To design an image retrieval system, we usually extract some visual features [3] from each image, and then compare the query image feature against the features of each database image and obtain the similarities to the query [4]. According to the similarities, the data base images will be ranked according to their similarities to the query on a descending order, so that the most similar ones will be ranked on the top ranks, while the non-similar ones will be on the bottom.

In this paper, we will investigate the issues of visual feature extraction and the similarity learning, and also the factorization of visual features, which are the most important issues in this system:

- Traditional **visual feature** extraction methods usually extract global features from an image, which is not robust to noise and rotation. Recently, bag-of-words methods have been proposed to represent image better than global features. For example, Park et al. [5], presents a simple and effective scene classification approach based on the incorporation of a multi-resolution representation into a bag-of-words model. Wang et al. [6] investigated using the bag-of-words strategy to classify the tissue types in medical imaging applications, and developed a novel algorithm that learns the vocabulary and weights jointly. Wang et al. [7] further improved the bag-of-words method by proposing an effective visual word weighting approach by analyzing each visual word's discriminating power by modeling the sub distance function. In [8], Wang et al. model the assignment of local descriptor as contribution functions, and then propose a new multiple assignment strategy.

- **Nonnegative matrix factorization (NMF)** has been used for the representation of visual feature representation for a long time. For example, Wang et al. [14] proposed a maximum correntropy criterion-based NMF method for gene expression data-based cancer clustering. Sawada et al. [15] paper presents new formulations and algorithms for multichannel extensions of non-negative matrix factorization. Wang et al. [16] proposed the multiple graph regularized nonnegative matrix factorization, by approximating the intrinsic manifold is by a linear combination of several graphs with different models and parameters inspired by ensemble manifold regularization. Wang et al. [17] proposed a novel data representation algorithm by integrating feature selection and graph regularization for NMF.

- Traditional **similarity** computation methods only consider a pair of images under consideration, and neglect the distribution of the database. Recently, contextual similarity learning has been proposed. For example, Jegou et al. [9] introduced the contextual dissimilarity measure by taking into account the local distribution of the vectors and iteratively estimates distance update terms in the spirit of Sinkhorn's scaling algorithm. Wang et al. [10] introduced the novel shape/object retrieval algorithm shortest path propagation (SSP), by explicitly finding the shortest path between the query and database image in the distance manifold of the database objects and learning a new distance measure between them. Bai et al. [11] provided a new perspective to the problem of Shape similarity and shape retrieval by considering the existing shapes as a group, and study their similarity measures to the query shape in a graph structure. Wang et al. [12] proposed a novel contextual protein-protein dissimilarity learning algorithm by regularizing an existing dissimilarity measure by considering the contextual information of the proteins. Moreover, Wang et al. [13] have developed the Multiple Graph regularized Ranking algorithm, by approximating the intrinsic manifold of image distribution by combining multiple initial graphs for the regularization.

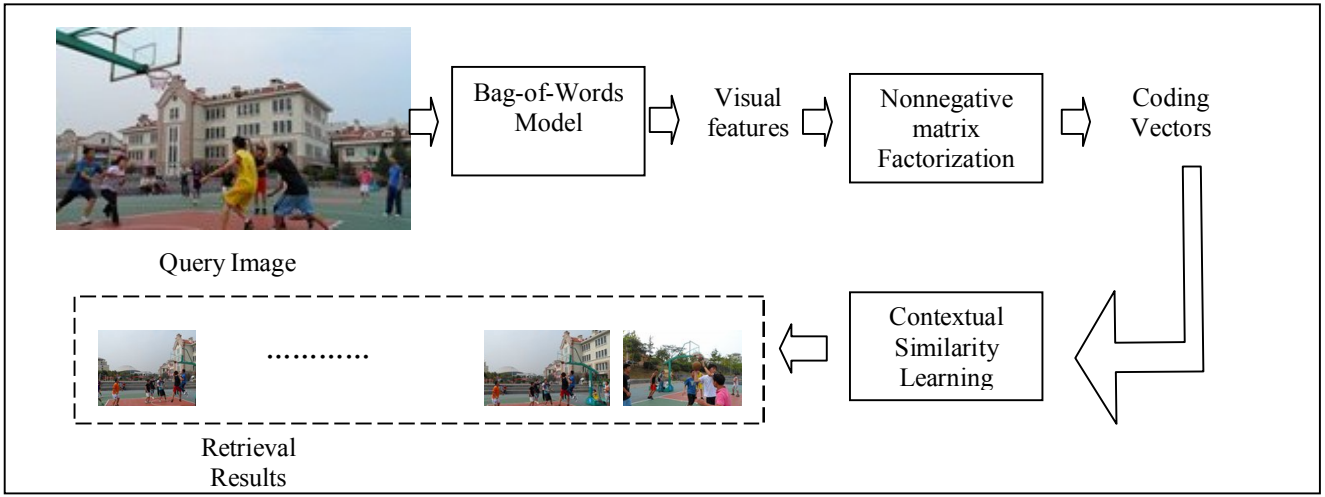

Figure 1. Framework of proposed system.

In this paper, we develop a novel image retrieval system, by integrating the bag-of-words model, NMF, and contextual similarity. The framework is shown in Figure 1. The paper is organized as follows: in section II, we introduce the proposed framework, and report the experiment results in section III, and conclude the paper in section IV.

## II. PROPOSED FRAMEWORK

In this section we will introduce the proposed Image retrieval system. The framework is shown in Figure 1.

### A. Bag-of-Words

The first phase is to extract the visual features from each image using bag-of-words medal. Each image will be split into many small image patches, and each patch will be treated as a "word" of an article, and each image as an "article". Then each patch will be quantized to a visual codebook and the quantization histogram is obtained as the visual feature. An example is shown in Figure 2.

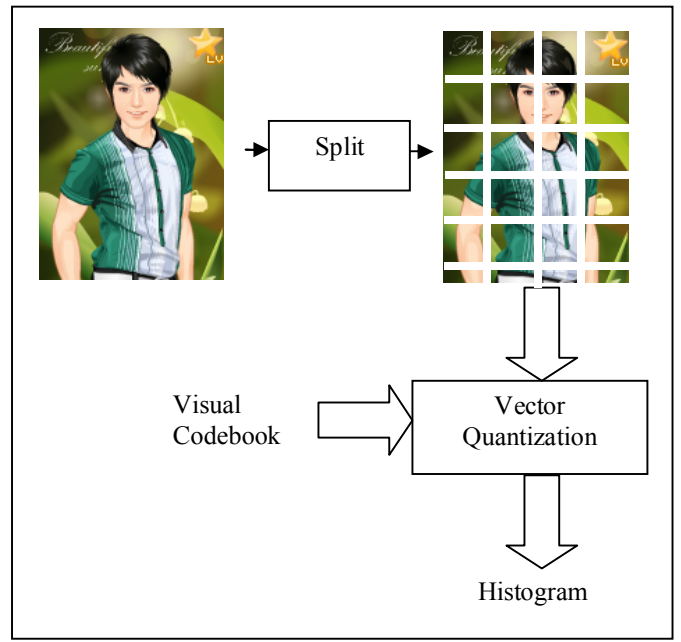

Figure 2. Bag-of-words based image representaion..

### B. Nonnegative Matrix Factorizations

The histograms of the training images will be organized as matrix and each column is a histogram. Then we could perform the NMF to the matrix. Given a basic matrix, each column will be characterized as the linear combination of some basic elements in the basic matrix, and the combination coefficients will be used as the new representation of the histograms. Thus the histogram matrix will be factorized as the product of a basic matrix and a coefficient matrix. We show this procedure in Figure 3.

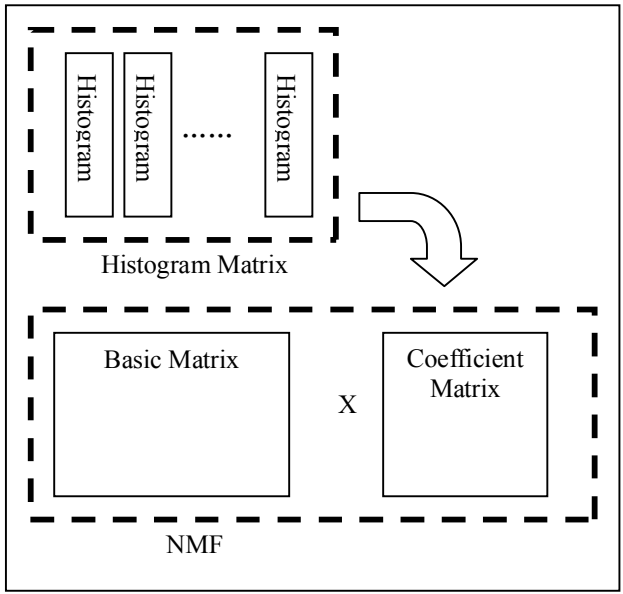

Figure 3. Histogram matrix factorization.

*C. Contextual Similarity Learning*

Using the representation of NMF of both query and database images, we could learn the contextual similarity by graph transduction introduced in [11]. Firstly a nearest neighbor graph will be constructed, and then the similarity will be transducted from the query to each database image. The transduction will be repeated for a few iterations and then used for ranking.

### III. EXPERIMENT

In this section, we will evaluate the performance of the proposed system on a large scale image database.

*A. Image database*

We collect an image database with 2,000 images. The images belong to 40 different classes, and each class has 50 images.

*B. Experiment protocal*

In this experiment, we employ the 10-fold cross-validation [18]. The entire database is split into 10 folds randomly, and each fold is used as a test set in turns while the rest 9 folds as training set. We use the ROC curve as the performance measure [19].

*C. Experiment Results*

The ROC curve is given in Figure 4. As we can see from the figures, the performances are satisfying.

### IV. CONLUSION

In this paper, we proposed a novel image retrieval system. In this system, we used the bag-of-words, NMF and contextual similarity. The performance is good. It's proven that by combining bag-of-words, NMF and contextual similarity technologies, we could have a effective image retrieval system.

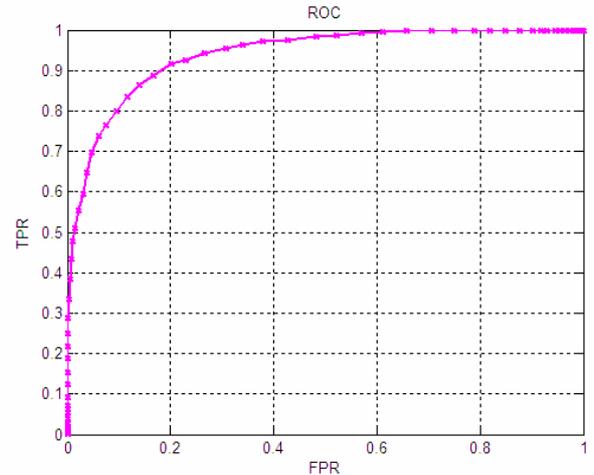

Figure 4. ROC of the proposed system..

ACKNOWLEDGMENT

This study was supported by the grant from State Key Laboratory for Novel Software Technology, Nanjing University, P.R. China (Grant No. KFKT2012B17).